\documentclass[times,twocolumn,final]{elsarticle}

\usepackage{framed,multirow}

\usepackage{amssymb}
\usepackage{latexsym}
\usepackage{amsmath,amssymb}
\usepackage{array}
\DeclareMathOperator{\argmax}{argmax}

\usepackage{url}
\usepackage{xcolor}
\usepackage{version}
\definecolor{modifs}{rgb}{0,0,0}
\definecolor{v2}{rgb}{0,0,0}
\usepackage{vmargin}
\setmarginsrb{50pt}{25pt}{50pt}{25pt}{50pt}{40pt}{100pt}{40pt}

\newcommand{\pathresspm}{.}
\graphicspath{{./}}

\newcommand{\spatch}[1]{\mathbf{#1}}
\newcommand{\eqsize}{\small}
\newcommand{\fig}{Fig.}
\newcommand{\eq}{Eq.}
\newcommand{\tab}{Tab.}
\newcommand{\sect}{Sec.}

\begin{document}

\begin{frontmatter}

\title{Multi-Scale Superpatch Matching using Dual Superpixel Descriptors}

\author[1]{R{\'e}mi {Giraud}}
\author[1]{Merlin {Boyer}}
\author[2]{Micha{\"e}l {Cl{\'e}ment}\corref{cor1}}
\cortext[cor1]{Corresponding author:
  Tel.: +33(0)540003552;}
\ead{michael.clement@labri.fr}

\address[1]{Bordeaux INP, Univ. Bordeaux, CNRS, IMS, UMR 5218, F-33400 Talence, France}
\address[2]{Bordeaux INP, Univ. Bordeaux, CNRS, LaBRI, UMR 5800, F-33400 Talence, France}

\begin{abstract}

Over-segmentation into superpixels is a very effective dimensionality reduction strategy, enabling fast dense image processing.
The main issue of this approach is the inherent irregularity of the image decomposition compared to standard hierarchical multi-resolution schemes, especially when searching for similar neighboring patterns.
Several works have attempted to overcome this issue by taking into account the region irregularity into their comparison model.
Nevertheless, they remain sub-optimal to provide robust and accurate superpixel neighborhood descriptors, since they only compute features within each region,
poorly capturing contour information at superpixel borders.
In this work, we address these limitations by introducing the dual superpatch, a novel superpixel neighborhood descriptor.
This structure contains features computed in reduced superpixel regions, as well as at the interfaces of multiple superpixels to explicitly capture contour structure information.
A fast multi-scale non-local matching framework is also introduced for the search of similar descriptors at different resolution levels in an image dataset.
The proposed dual superpatch enables to more accurately capture similar structured patterns at different scales, and we demonstrate the robustness and performance of this new strategy on matching and supervised labeling applications.
\end{abstract}

\end{frontmatter}


\section{Introduction}

In many computer vision related applications, such as image classification and segmentation, there is a important need for fully automated results.
{\color{v2}To this end, a commonly employed strategy is to take inspiration from other data,
in a supervised way when ground truth annotations are available.
In this context, non-local methods have provided accurate results on various applications.
In these methods,
image regions are independently considered, generally using a square patch defined for each pixel, capturing a local pattern \cite{buades2005}.
For exemplar-based classification or segmentation,
matching algorithms are usually used to search for similar patterns in the available data, to then transfer the associated information,
at the pixel or image scale after a global decision process.}

{\color{v2}
This search for similar patterns is generally performed for each image patch, and
fast matching algorithms have been developed
\emph{e.g.}, PatchMatch~\cite{barnes2009},
TreeCANN~\cite{olonetsky2012}
or FLANN \cite{muja2014scalable},
to efficiently exploit a large number of example images in a reduced computational time.}
To find similar content with such matching algorithms,
it is common to extract features from dense image patches or local interest points.
These features are usually designed to be robust to transformations such as scaling, viewpoint or illumination changes.
Standard descriptors include features based on gradient information such as SIFT~\cite{lowe2004} or HoG~\cite{dalal2005,lsvm-pami}, or binary patterns, \emph{e.g.}, BRIEF~\cite{calonder2010brief}.

In the last few years,
convolutional neural networks have also been able to extract particularly relevant features, and have shown promising results in many applications related to image processing~\cite{lecun2015deep}.
However, {\color{v2}although some recent architectures such as U-Net may learn from relatively small datasets \cite{ronneberger2015u}}, these methods rely on costly supervised learning strategies, and often require very large annotated datasets to be trained efficiently.
Besides, the results of these models are generally difficult to interpret, and can be very sensitive to small perturbations of
their inputs~\cite{moosavidezfooli2016universal}.
In many cases, such as medical imaging,
these drawbacks
can limit the potential of automated
processing pipelines.
Therefore, there is still an important need for methods that can perform without learning steps, as well as with limited training data and computing power.

In this context of fast image search and matching requirements, many works have first focused on hierarchical approaches using prior image over-segmentation
into regular grids, \emph{e.g.}, \cite{lazebnik2006beyond}.
To go further, other methods proposed to group similar pixels into connected components of homogeneous colors called superpixels,
drastically reducing the number of elements to process
while preserving contours and spatial structure~\cite{stutz2018}.
Therefore, a process applied at such over-segmentation scale can be close to the optimal pixel-wise result.
{\color{v2}
Several works have used
superpixels in non-local frameworks, \emph{e.g.}, \cite{gould2008,tighe2010},
or in unsupervised learning-based superpixel matching approaches using random forests
\cite{conze2017hierarchical,kanavati2017supervoxel}.
Nevertheless, the geometrical irregularity of such decompositions~\cite{giraud2017_gef}
{\color{v2}(\emph{i.e.}, in terms of shape, adjacency or contour smoothness)}
can become an issue, since
neighborhood information is crucial to compute accurate matches in terms of context.
}

Other approaches have attempted to use superpixel neighboring information, \emph{e.g.}, \cite{pei2014,sawhney2014}.
Among them, the SuperPatchMatch (SPM) framework~\cite{giraud2017_spm} partially addresses this issue
with a superpixel neighborhood structure called superpatch and a dedicated 
metric to
compare two structures having different geometry and number of elements.
However, SPM remains sub-optimal
in terms of computational complexity and matching accuracy.
The framework enables the search for matches at the same superpatch scale and it only computes features within each superpixel region, poorly capturing contour information.
{\color{v2}Several works indeed highlighted the need for accurate superpixel-wise features~\cite{neubert2015benchmarking,zhang2018consistent,tilquin2018robust},
while most image descriptors are locally computed on a regular square neighborhood.}

\subsection*{Contributions}

{\color{v2}
In this paper, we address the limitations of previous non-local methods only focusing on intra-region information within superpixels or superpixel neighborhoods \cite{giraud2017_spm}},
by introducing a novel dual superpixel neighborhood descriptor called dual superpatch (DSP), containing two independent descriptor sets (see \sect~\ref{sec:dual}).

First, intra-superpixel features capture color or texture information within cropped superpixel regions in order to avoid influence of pixel contours or inaccurate superpixel borders (see \sect~\ref{subsec:intra}).
Then, to capture structure information,
{\color{modifs}
for instance in terms of contour orientations,
}%
we extract a relatively regular grid of specific descriptors at superpixel interfaces (see \sect~\ref{subsec:sid}).
To efficiently compare such irregular dual descriptors, having different geometry and number of elements,
we also propose new distances and optimizations, significantly reducing the computational complexity.

The SuperPatchMatch (SPM) search algorithm \cite{giraud2017_spm}
with our accurate dual superpatch (DSPM),
performs more relevant superpixel matching.
To go further, we also extend DSPM to the search of matches at multiple scales (see \sect~\ref{sec:multiscale}),
and propose a framework to perform automatic labeling using exemplar-based images with ground truth labels.
In our framework, the comparison of DSP at different scales can be easily performed since we consider reduced spatial information, \emph{i.e.}, sets of barycenter positions.
This way, we are able to match similar objects at different sizes in heterogeneous datasets.

Finally, to show the robustness of our framework, especially compared to \cite{giraud2017_spm}, we consider several matching and exemplar-based labeling experiments on a standard face dataset \cite{huang2007db} (see \sect~\ref{sec:experiments}).

\section{The SuperPatchMatch Framework}

In this section, we first recall the SuperPatchMatch (SPM) framework initially introduced by~\cite{giraud2017_spm}, which constitutes the basis of our approach.

\subsection{The SuperPatch Structure}

\begin{figure}[t!]
    \centering
    \includegraphics[width=0.475\textwidth]{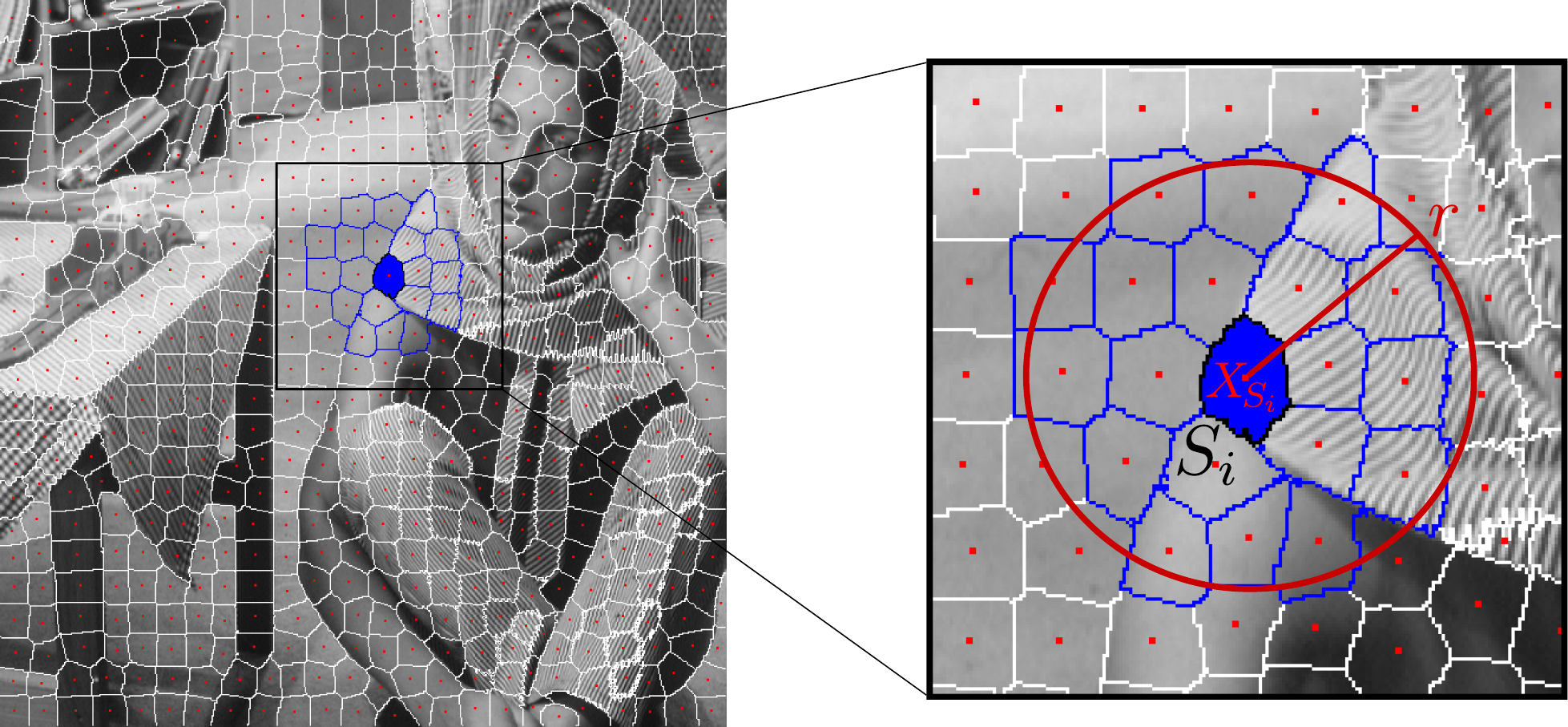}
    \caption{Superpatch definition. For a superpixel $S_i$ (filled blue), neighboring superpixels (blue contours) having their barycenters (red dot) into a radius $r$, centered on $X_{S_i}$ the barycenter of $S_i$, are part of the superpatch $\mathbf{S_i}$.}
    \label{fig:spp}
\end{figure}

To generalize standard patch-based frameworks
to irregular image decompositions,
\cite{giraud2017_spm} proposed the \textit{superpatch} structure.
As for square patches of pixels defined around each pixel,
a superpatch $\spatch{S_i}$, associated to a superpixel $S_i$,
contains the adjacent neighbors of a superpixel $S_i$ with
respect to a fixed radius $r$.
The proximity is simply computed according to the superpixel spatial barycenters such that:
{\eqsize
\begin{equation}
\label{spp}
\spatch{S_i}=\{ S_{i'}, \textrm{ such that }  {||X_{S_i}-X_{S_{i'}}||}_2 \leq r \},
\end{equation}
}%
where $X_{S_i}=[x_{S_i},y_{S_i}]$ and $X_{S_{i'}}=[x_{S_{i'}},y_{S_{i'}}]$ respectively denote the spatial barycenters of superpixels $S_i$ and $S_{i'}$.
This way, the superpatch structure only includes the most significantly neighboring superpixels, using reduced spatial information.
In \fig~\ref{fig:spp}, we show a superpatch example, defined for a superpixel $S_i$, containing its adjacent superpixels to provide a superpixel neighborhood descriptor.

\subsection{SuperPatch Comparison Distance}

A comparison distance is also proposed in \cite{giraud2017_spm} to measure the similarity
between two superpatches.
The main issue to design such distance is that the two structures are very likely to have different geometry and number of elements.
Hence, there is no one-to-one association between superpixels, contrary to pixels within regular patches.
To preserve the ability to compare patterns, the spatiality must be taken into account, and \cite{giraud2017_spm} proposed to simply consider the proximity of superpixel barycenters after registration on the central superpixels.
In the following, we consider two superpatches $\spatch{S_i}$ and $\spatch{S_j}$, for instance in two images $A$ and $B$.
A weight $w(X_{S_{i'}},X_{S_{j'}}) =\exp(-\|X_{S_{j'}} - (X_{S_{i'}} - (X_{S_i}-X_{S_j}))\|_2^2/\sigma_1^2)$,
measures
the relative displacement between the registered barycenters $X_{S_{i'}}$ and $X_{S_{j'}}$ of superpixels $S_{i'}\in\spatch{S_i}$ and $S_{j'}\in\spatch{S_j}$, with respect to central superpixels $S_i$ and $S_j$, and $\sigma_1$ is a scaling parameter set to $1/2\sqrt{|I|/K}$, with $|I|$ and $K$ the respective number of pixels and superpixels in the image.
This way a superpixel $S_{i'}$ only compares to the closest ones in $\spatch{S_j}$.

The distance $D$ between two superpatches $\spatch{S_i}$ and $\spatch{S_j}$ is finally defined as:
{\small
\begin{equation}
D(\spatch{S_i},\spatch{S_j})=\frac{\sum\limits_{S_{i'}\in \spatch{S_i}}\sum\limits_{S_{j'}\in \spatch{S_j}}^{} w(X_{S_{i'}},X_{S_{j'}})w_s(X_{S_{i'}})w_s(X_{S_{j'}}) d(F_{S_{i'}},F_{S_{j'}})}{\sum\limits_{S_{i'}\in \spatch{S_i}}\sum\limits_{S_{j'}\in \spatch{S_j}}^{} w(X_{S_{i'}},X_{S_{j'}})w_s(X_{S_{i'}})w_s(X_{S_{j'}})},\label{dist3_spp}
\end{equation}}%
\noindent
where
$w_s(X_{S_{i'}})$ also weights the influence of $S_{i'}$ according to
its spatial distance to $S_i$ such that
$w_s(X_{S_{i'}})=\exp(-\|X_{S_{i'}} - X_{S_i}\|_2^2/(2r^2))$, and
$d$ is the distance between the superpixel features $F_{S_{i'}}$ and
$F_{S_{j'}}$.
{\color{modifs}
Note that any distance $d$ and feature $F$ can be considered in Eq. \eqref{dist3_spp}.
}%

\begin{figure}[t!]
    \centering
    \includegraphics[width=0.45\textwidth]{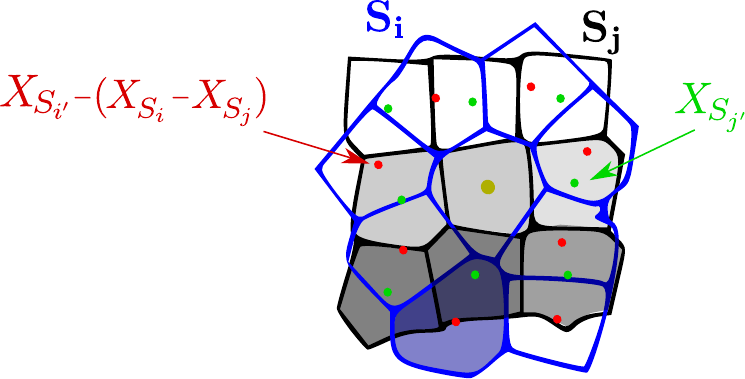} \\[-1ex]
    \caption{Comparison process of two superpatches $\spatch{S_i}$ and $\spatch{S_j}$
    from \eq~\eqref{dist3_spp}.
    Superpatches are registered according to the barycenters $X_{S_i}$ and $X_{S_j}$ of their central superpixel.
    The weights $w$ in \eq~\eqref{dist3_spp} favor the comparison to closest superpixels after registration and the weight values corresponding to the bottom superpixel of $\spatch{S_i}$
    {\color{modifs}
    (filled blue)
    }%
    are represented within each compared superpixel of $\spatch{S_j}$ (darker meaning a higher $w$ weight).
    }
    \label{fig:spm_recalage}
\end{figure}

{\color{modifs}
The comparison process between two superpatches having different number of element and geometry
}%
is illustrated in \fig~\ref{fig:spm_recalage}.
The weight $w$ in \eq~\eqref{dist3_spp} weights the feature distance $d$ between a superpixel in $\spatch{S_i}$ and a superpixel in $\spatch{S_j}$
after registration on their central barycenters. 
In this Figure, the weights $w$ corresponding to the bottom superpixel of $\spatch{S_i}$ are represented within each superpixel of $\spatch{S_j}$.

\subsection{SuperPatchMatch Correspondence Algorithm}

Non-local methods have soon highlighted the need for fast patch-based matching algorithms to perform the search of correspondences within large areas, \emph{e.g.}, library of example images.
A significant breakthrough has been obtained with PatchMatch (PM) \cite{barnes2009},
a fast partly random matching algorithm,
providing for each patch of an image $A$, a match in an image $B$.
PM has very interesting properties such as no requirements for learning or preprocessing steps,
and its complexity only depends on the size of the image to process $A$,
enabling to search for matches in a large set of example images.

The PM algorithm starts from random associations and iteratively refines them with a sequential processing of all image patches.
The refinement process is mainly based on the fast propagation of good matches from spatially adjacent neighbors. Large regions are indeed very likely to correspond between images.
According to the scan order, which is reversed at each iteration, the shifted correspondences of two spatially adjacent patches are considered as new candidates.
Also, random patches are tested near the current best match in $B$.
Note that the PM process being partly random, several processes carried out in parallel can potentially provide
different matches.

\begin{figure}[t!]
    \centering
    \includegraphics[width=0.485\textwidth]{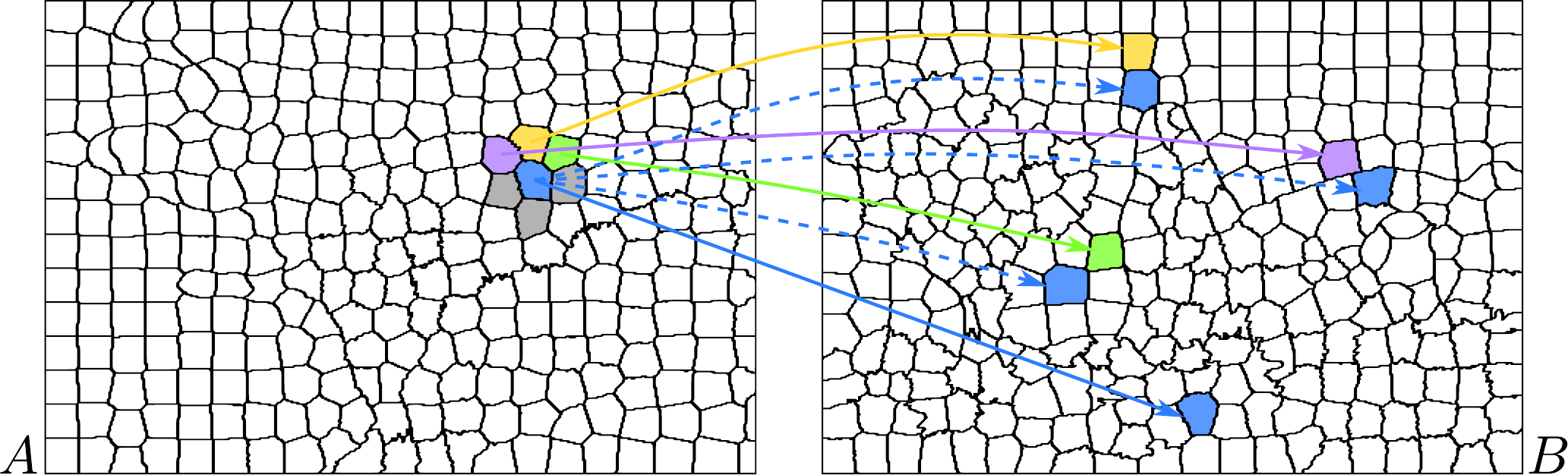}
    \caption{Core of the SPM algorithm.
    Full lines indicate current best matches.
    The direct adjacent neighbors of
    {\color{modifs}
    the
    }%
    blue superpixel are considered to propose new match candidates (dotted lines).
    The relative orientations between superpixels in $A$ tend to be respected in $B$, \emph{e.g.}, the yellow superpixel remains on top of the blue one in both images.
    Remaining adjacent neighbors (gray) that have not yet being processed during this iteration will be considered at the next one, processing superpixels in the reverse order.
    } 
    \label{fig:spm}
\end{figure}

The SuperPatchMatch (SPM) method generalizes PM to superpatches \cite{giraud2017_spm}, to provide a fast matching algorithm of superpixels.
The method mainly requires to adapt the propagation of matches based on adjacent
neighbors since there is no more consistent geometry between adjacent superpixels, contrary to the standard pixel grid case.
The propagation step of SPM is illustrated in \fig~\ref{fig:spm} in the case of two images $A$ and $B$.
The adjacent neighbors are considered to lead to new matches while respecting the relative orientation between superpixels in $A$ and $B$, to favor the matching of larger regions.

\subsection*{Limitations}

Nevertheless, the default SPM distance
{\color{modifs}in \eq~\eqref{dist3_spp}}
has a quadratic complexity, \emph{i.e.}, each superpixel of a superpatch is compared to all the ones of the other superpatch, and may result in important computational time.
Besides, this framework only considers intra-superpixel descriptors.
Although, they may be sufficient to capture information within regions,
the superpatch does not explicitly focus on capturing contours or gradient information.
This may be an issue since such information generally lies at the border of a superpixel,
and can be shared between two regions, reducing the relevance of the descriptor.
Finally, the SPM framework does not consider multi-scale information that would allow to capture objects of different sizes.

We address all these issues in the following sections with the proposed multi-scale superpatch matching framework that uses new dual superpixel descriptors.

\section{\label{sec:dual}Dual Superpixel Descriptors}

\begin{figure}[t!]
    \centering
    \includegraphics[width=0.4\textwidth,height=0.26\textwidth]{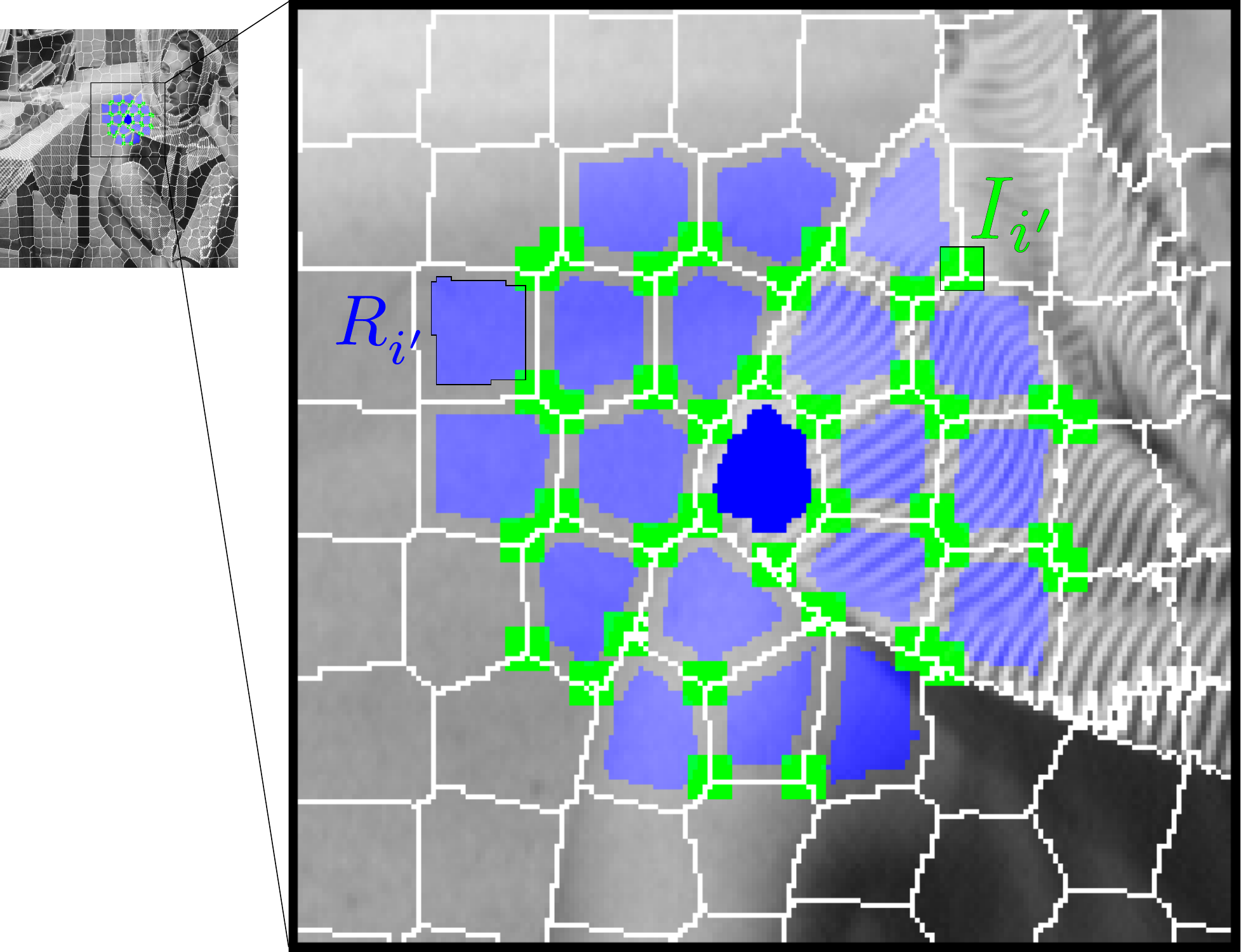}
    \caption{Illustration of the dual superpatch (DSP) descriptor.
    Intra-region information ($R_{i'}$) within each superpixel $S_{i'}$ with an offset of $\beta=3$ pixels from the border (blue regions) are considered along with information at the superpixel interfaces $I_{i'}$ (green squares) within the same $r$ radius.
}
    \label{fig:spp_dual}
\end{figure}

{\color{v2}
In this section, we propose an approach to relevantly extract superpixel neighborhood descriptors.
We introduce a dual descriptor to efficiently capture around a superpixel,
both feature region content  and
contour information, potentially lying at their borders.}
Features are computed within superpixel regions and additional contour features are extracted at multiple interfaces between adjacent superpixels.
The proposed dual descriptor of superpixel neighborhood is called \textit{Dual SuperPatch} (DSP), is denoted
$\spatch{\bar{S}_i}$ for a superpixel $S_i$,
and is represented in \fig~\ref{fig:spp_dual}
on the same decomposition example used in \fig~\ref{fig:spp}.
In the following, we present the extraction approach for
region (R), \emph{i.e.}, intra-superpixel,
and interfaces (I) descriptors,
and we propose a general framework to compare different DSP.

\subsection{Superpixel Region Descriptors\label{subsec:intra}}

{\color{v2}
The superpatch formulation of~\cite{giraud2017_spm}
considers features computed on
each whole superpixel region contained into the neighborhood structure.}
However, superpixels tend to capture homogeneous regions,
so pixels at thin contours can be arbitrarily associated to superpixels, leading to altered descriptors.
A reduced block area or spatial weighting from the superpixel barycenter could not be applied
since these approaches do not guarantee to relevantly extract
information when superpixels have very irregular shapes~{\color{modifs}\cite{neubert2015benchmarking}.}%

In this work, we propose to consider the superpixel region information with an offset of $\beta$ pixels to its borders.
This way, we take into account almost all the region, while being robust to inaccurate superpixel borders or contour information,
that will be considered
in another specific interface descriptors within our
DSP (see \sect~\ref{subsec:sid}).
{\color{modifs}
In \fig~\ref{fig:spp_dual}, the considered regions  $R_{i'}$ for superpixels $S_{i'}$ are represented in blue.
}%
To each region $R_{i'}$, feature $F_{R_{i'}}$ and spatial $X_{R_{i'}}$ information are considered, so a dual superpatch contains a set of tuples $\spatch{R_i}=\{F_{R_{i'}},X_{R_{i'}}\}$.

\begin{figure}[t]
\centering
{\small
\begin{tabular}{@{\hspace{0mm}}c@{\hspace{3mm}}c@{\hspace{0mm}}}
\includegraphics[width=0.225\textwidth,height=0.13\textwidth]{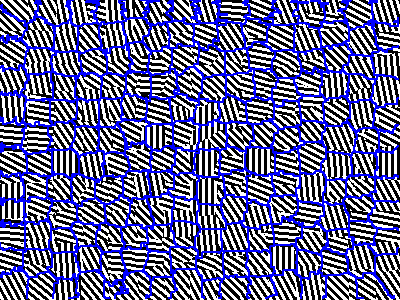}&
\includegraphics[width=0.225\textwidth,height=0.13\textwidth]{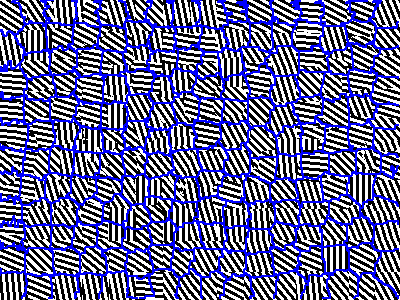}\\ 
\end{tabular}
}
\caption{Two synthetic images containing $16$ oriented textures and decomposed into superpixels with \cite{giraud2019_nnsc}.}
\label{fig:expe_texture}
\end{figure}

To demonstrate the issue of considering the whole superpixel region to extract features, we consider two decompositions of images containing regions with $16$ different oriented textures (see \fig~\ref{fig:expe_texture}).
The superpatch radius is set to $r=0$ to only consider intra-region information, where HoG descriptors \cite{dalal2005} are computed.
{\color{v2}
For each superpixel of the left image, we compute in an exhaustive manner its closest match in terms of superpixel content in the right image.
If the texture are similar, we consider the matching as accurate (1), otherwise inaccurate (0).
We report in \tab~\ref{table:expe_texture} the average matching accuracy on all superpixels of the left image,
according to different $\beta$ values and several levels of Gaussian noise applied to both images
after decomposition.
We perform this evaluation using the decompositions obtained with a texture-aware method \cite{giraud2019_nnsc}, and also the ground truth ones, perfectly capturing texture changes.
This experiment highlights the need to restrict the area to extract superpixel information since superpixel decompositions may be not perfectly accurate.
Moreover, even on perfectly fitting decompositions,
the inaccurate gradient information lying on superpixel borders is captured with $\beta=0$ and may impact the results
with a noise variance superior to 50, while we do not take into account texture in other superpixels with $\beta>0$.}

\begin{table}[t]
\caption{Impact of border offset $\beta$ for intra-region descriptor extraction.
Texture matching results between images in \fig~\ref{fig:expe_texture} 
for different
values of $\beta$ and Gaussian noise levels
using both superpixel decompositions from \cite{giraud2019_nnsc} and ground truth ones.
Best and second results are respectively bold and underlined.}
\begin{center}
\renewcommand{\arraystretch}{1.15}
{\scriptsize
\begin{tabular}{@{\hspace{1mm}}c@{\hspace{4mm}}c@{\hspace{2mm}}c@{\hspace{2mm}}c@{\hspace{2mm}}c@{\hspace{5mm}}c@{\hspace{2mm}}c@{\hspace{2mm}}c@{\hspace{0mm}}c@{\hspace{0mm}}} \cline{2-9}
& \multicolumn{4}{@{\hspace{0mm}}c@{\hspace{3mm}}}{Superpixel decompositions \text{ }\text{ }\text{ }} &    
  \multicolumn{4}{@{\hspace{0mm}}c@{\hspace{2mm}}}{Ground truth decompositions}   \\  \cline{1-9}
Noise variance & 0 & 50 & 100 & 125 & 0 & 50 & 100 & 125  \\ \hline
$\beta=0$    & $0.526$  & $0.634$  & $0.516$  & $0.366$ & $\mathbf{1.000}$  & $\mathbf{1.000}$  & $0.980$ & $0.907$ \\
$\beta=1$   & $0.616$  & $0.654$  & $\mathbf{0.558}$ & $0.419$  & $\mathbf{1.000}$  &$\mathbf{1.000}$  & $\mathbf{0.993}$ &  $\mathbf{0.913}$ \\
$\beta=2$  & $\underline{0.679}$  & $\underline{0.665}$  & $\underline{0.558}$  & $\mathbf{0.482}$ & $\mathbf{1.000}$  & $\mathbf{1.000}$  & $\underline{0.987}$  & $\underline{0.913}$ \\
$\beta=3$  & $\mathbf{0.711}$  & $\mathbf{0.675}$  & $0.521$  & $\underline{0.482}$ & $\mathbf{1.000}$  & $\mathbf{1.000}$  & $0.953$  & $0.900$ \\
\hline
\end{tabular}
}
\end{center}
\label{table:expe_texture}
\end{table}

\begin{figure*}[t]
\centering
{\scriptsize
\begin{tabular}{@{\hspace{0mm}}c@{\hspace{8mm}}c@{\hspace{0mm}}}
\includegraphics[width=0.44\textwidth,height=0.11\textwidth]{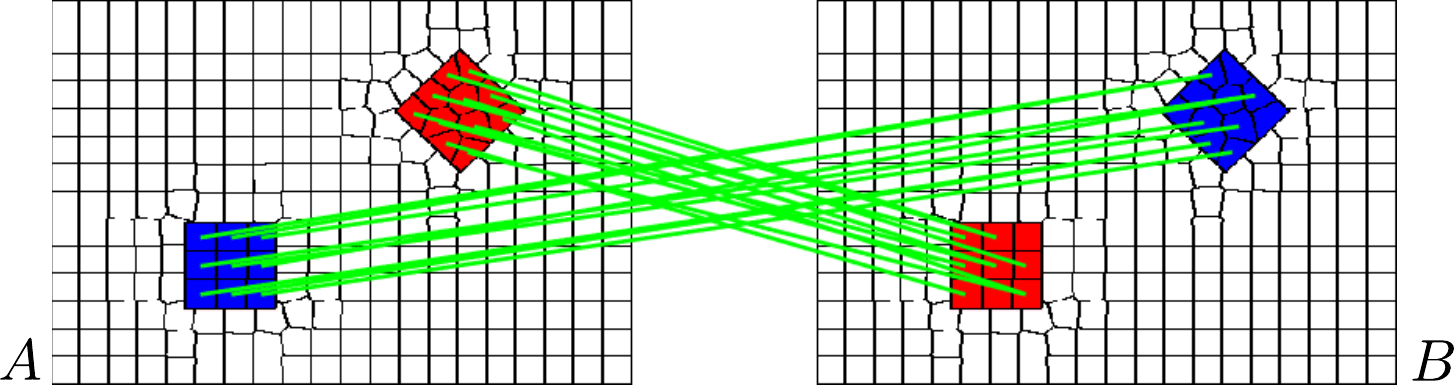}&
\includegraphics[width=0.44\textwidth,height=0.11\textwidth]{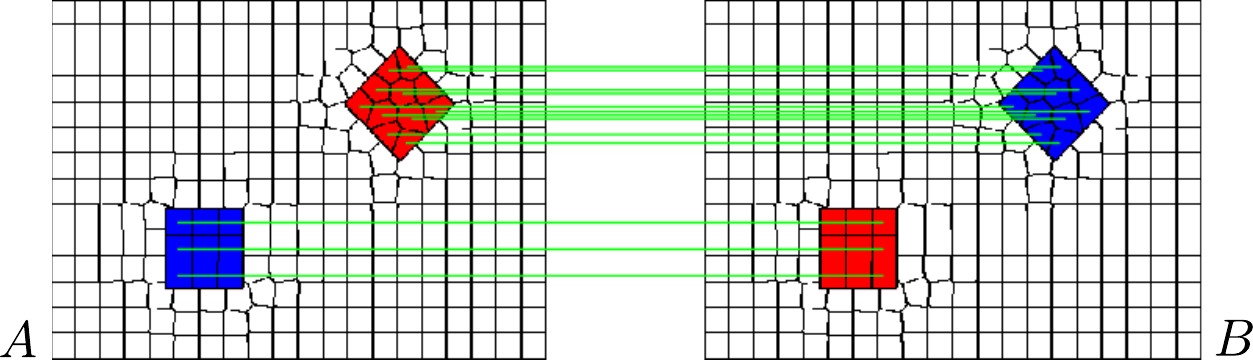}\\[1ex]
\multicolumn{1}{@{\hspace{0mm}}c@{\hspace{8mm}}}{(a) Matching using only color information at intra-superpixel regions ($\alpha=1$)}&
\multicolumn{1}{@{\hspace{0mm}}c@{\hspace{0mm}}}{(b) Matching using only contour features at superpixel interfaces ($\alpha=0$)} \\ 
\end{tabular}
}
\caption{Impact of the $\alpha$ parameter in the DSP distance \eqref{full_model}.
(a) Only region descriptors $R_{i'}$ are used ($\alpha=1$), with average color information.
This result corresponds to the one obtained using \cite{giraud2017_spm}.
(b) Only interface HoG~\cite{dalal2005} descriptors $I_{i'}$ are used ($\alpha=0$) and enable to capture structure information.
Radius $r$ is set to $25$, \emph{i.e.}, approximately capturing the first adjacency ring.}
\label{fig:sid_example}
\end{figure*}

\smallskip

\textit{Fast comparison distance.}
The comparison between two sets of
superpixel region descriptors can be performed in a more computationally efficient manner than with 
\eq~\eqref{dist3_spp}.
We propose to only select one superpixel $S_{j'}\in \spatch{S_j}$ to compare for each superpixel $S_{i'}\in \spatch{S_i}$.
{\color{v2}
To do so, the superpixel barycenter $X_{S_{i'}}$ is first registered by the displacement between central superpixels $S_{i}$ and $S_{j}$, and we denote this new position by $X_{S^j(i')}$ which is computed such that
$X_{S^j(i')} = X_{S_{i'}} - (X_{S_i} - X_{S_j})$.
In \fig~\ref{fig:spm_recalage}, these correspond to the red superpixel barycenters.
Then, we project the registered barycenters on the decomposition of the image from where $\spatch{S_j}$ is extracted. 
In \fig~\ref{fig:spm_recalage}, the black superpixel containing a red dot $X_{S^j(i')}$ would be selected to compare to superpixel $S_{i'}$ of region $R_{i'}$.
This corresponding superpixel containing $X_{S^j(i')}$ in the compared image, is denoted $S^j{(i')}$, and its associated intra-region is denoted $R^j{(i')}$.}
This way, we significantly reduce the distance complexity,
while potentially increasing the comparison accuracy (see Sec. \ref{sec:experiments}).
The comparison between two region descriptors $\spatch{R_i}$ and $\spatch{R_j}$ is defined using barycenter projections such that:
\begin{equation}
 d_{p}(\spatch{R_i},\spatch{R_j})=
 \frac{\sum\limits_{R_{i'}\in \spatch{R_i}}w_s\left(X_{R_{i'}}\right)d\left(F_{R_{i'}},F_{R^j{(i')}}\right)}{\sum\limits_{R_{i'}\in \spatch{R_i}}w_s\left(X_{R_{i'}}\right)} .
 \label{dist_R1}
\end{equation}
\noindent
Note that barycenters falling outside the image limits are projected to select the closest superpixel on the image boundary.

{\color{modifs}
A similar projected
}%
distance was suggested in \cite{giraud2017_spm}, in a non-symmetric formulation.
In our dual superpatch comparison model,
{\color{modifs}
we consider a symmetric projected distance $D_p$
}%
on intra-region descriptors defined as:
\begin{equation}
 D_{p}(\spatch{R_i},\spatch{R_j})= \frac{1}{2}\left(d_p\left(\spatch{R_i},\spatch{R_j}\right)+d_p\left(\spatch{R_j},\spatch{R_i}\right)\right) .
 \label{dist_R}
\end{equation}

\subsection{Superpixel Interface Descriptors\label{subsec:sid}}

To efficiently capture image contour information,
we propose to also consider specific descriptors at superpixel interfaces.
These can be easily extracted with a low complexity by considering the presence of at least three superpixels in a $3{\times}3$ pixels neighborhood.
To avoid over detection, a larger area can be neglected after selection of an interface point.
This way, we directly obtain a relatively regular grid of potential interest points in terms of contours,
without introducing further scaling or thresholding parameters.
In \fig~\ref{fig:spp_dual}, these interface regions denoted $I_{i'}$ are represented as green squares.
On these regions, specific contour descriptors can be computed,
\emph{e.g.}, HoG
\cite{dalal2005}.

\smallskip

\textit{Acceleration of quadratic distance.}
Since interface regions do not provide a dense decomposition of the image domain, the distance \eq~\eqref{dist_R1} using projections cannot be used to fastly compare two sets of interface descriptors.
{\color{v2}A quadratic one-to-many distance such as \eq~\eqref{dist3_spp} could be used, but at the expense of important computational cost.
To address this issue, we propose a one-to-one association for each interface descriptor $I_{i'}$.
Each $I_{i'}$ is only compared to the spatially closest one $I_{j'}$ in the other dual superpatch.
This way, the framework only requires exhaustive spatial distances between interface barycenters.
The distance is computed as for Eq. \eqref{dist_R1}, where $I^j(i')$, the selected interface descriptor in $\mathbf{I_j}$ for $I_{i'}$
is defined as $I^j(i')=\underset{I_{j'}}{argmin}{\|X_{I_{i'}}-X_{I_{j'}}\|_2}$.
Finally, as for Eq. \eqref{dist_R}, the distance is also computed from $\mathbf{I_j}$ to  $\mathbf{I_i}$ to obtain a symmetric distance.}

\subsection{General Dual SuperPatch Comparison Framework}

Our dual superpatch (DSP) $\spatch{\bar{S}_i}$,
{\color{modifs}
for a superpixel $S_i$,
}%
is described by a set of
intra-superpixel regions $\spatch{R_i}=\{F_{R_{i'}},X_{R_{i'}}\}$
and superpixel interfaces $\spatch{I_i}=\{F_{I_{i'}},X_{I_{i'}}\}$
descriptors such that
$\spatch{\bar{S}_i}=[\spatch{R_i},\spatch{I_i}]$.
Note that $\spatch{R_i}$ and $\spatch{I_i}$ can have a different number of elements.
To relevantly measure the similarity of two DSP
$\spatch{\bar{S}_i}$ and $\spatch{\bar{S}_j}$
having different geometry and number of elements,
we propose the following general DSP comparison distance:
\begin{equation}
    D(\spatch{\bar{S}_i},\spatch{\bar{S}_j})
        = \alpha \hspace{0.05cm} D_{p}(\spatch{R_i},\spatch{R_j}) +
        (1-\alpha) \hspace{0.05cm} {D_{p}(\spatch{I_i},\spatch{I_j})},  \label{full_model}
\end{equation}
\noindent
with $D_{p}$ the fast distance on descriptors, using barycenter projections  \eq~\eqref{dist_R} for intra-region $R$, and selection of closest descriptor for interfaces $I$.
and $\alpha \in [0, 1]$ a setting parameter.
Note that any feature can be considered in $F_{R_{i'}}$ or $F_{I_{i'}}$.
Therefore, $\alpha$ can have an intuitive tuning using
the same or normalized
descriptors for both $R$ and $I$.

In \fig~\ref{fig:sid_example},
we show matching results obtained with our generalized model using average RGB color as intra-region features $F_{R_{i'}}$ (\fig~\ref{fig:sid_example}(a), $\alpha = 1$) and HoG descriptors as interface features $F_{I_{I'}}$ (\fig~\ref{fig:sid_example}(b), $\alpha = 0$).
Hence, we highlight the general aspect of our framework
that allows to either focus on intra-region (\fig~\ref{fig:sid_example}(a)) or interface information (\fig~\ref{fig:sid_example}(b)).
In \sect~\ref{sec:experiments}, we further demonstrate the performances obtained using these complementary descriptors.

\section{Multi-Scale Dual SuperPatchMatch\label{sec:multiscale}}

In this section, we propose to extend the SPM framework with our dual descriptor (DSPM), to perform the search of DSP at multiple scales.
We first show how to compare two DSP of different sizes, then we propose a multi-scale fusion strategy.

\subsection{Dual SuperPatch Rescale\label{subsec:rescale}}

In \sect~\ref{sec:dual}, we showed how to compare two DSP extracted with the same radius size $r$.
Nevertheless, the proposed distance  \eq~\eqref{full_model} can easily adapt to DSP of different sizes,
since the spatial information is only measured by barycenter positions denoted with $X$.
We consider two DSP $\spatch{\bar{S}_i}$ and $\spatch{\bar{S}_j}$,
with different DSP extraction radius $r^i$ and $r^j$ in \eq~\eqref{spp}.
To compare them, all spatial information contained in $\spatch{\bar{S}_j}$ can be adjusted according to the ratio between the radiuses, such that:
\begin{equation}
\spatch{\bar{S}_j}=
{\color{modifs}
[\spatch{R_j},\spatch{I_j}]
}%
=
\left[\left\{\left(F_{R_{j'}},X_{R_{j'}}\frac{r^i}{r^j}\right)\right\},\left\{\left(F_{I_{j'}},X_{I_{j'}}\frac{r^i}{r^j}\right)\right\}\right] . \label{rescale}
\end{equation}
This way,
similar DSP can be searched at various scales, \emph{e.g.}, in example images.
Note that features $F_{R_{j'}}$ and $F_{I_{j'}}$ remain unchanged by this scaling transformation.

\subsection{Multi-Scale Exemplar-based Framework}
\label{subsec:fusion}

Most non-local methods %
perform a search for similar content
in a heterogeneous dataset,
with no prior information on the targeted object size.
In \cite{giraud2017_spm}, no multi-scale strategy is proposed
since it considers an exemplar-based labeling experiment
on linearly registered images \cite{huang2007db}.

Here, we introduce a generalized multi-scale exemplar-based framework allowing to search for similar DSP of different radiuses, in order to capture objects of different sizes.
The proposed DSP structure indeed enables to perform a simple automatic rescale, presented in \sect~\ref{subsec:rescale}.
Hence, multiple DSP sizes can be considered
in a set of examples images $\mathbf{B}$.
A set $\mathbf{r^B}=\{r^B\}$
is considered for
setting the DSP radius in
$\mathbf{B}$.

\begin{figure}[t]
\centering\includegraphics[width=0.8\linewidth,height=0.1\textwidth]{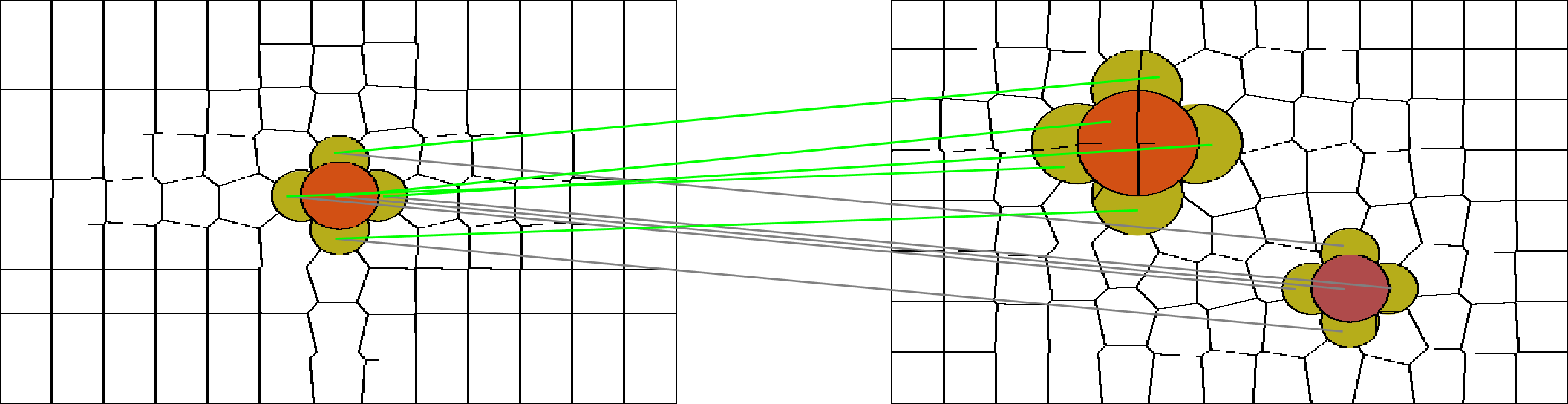}
\caption{Example of matching result without (gray lines)
and with (green lines) a multi-scale search for matches. See text for more details.
} 
\label{fig:expe_ms}
\end{figure}

In \cite{giraud2017_spm}, a supervised labeling framework based on the non-local means algorithm \cite{buades2005} was introduced to merge the information of multiple superpatch matches computed in a library of example images for an image $A$ to process.
A label map $L_m^{r^B}(S_i)$ is computed for a superpixel $S_i$, for all $M$ different labels $m$, such that: \vspace{-0.1cm}
{\color{modifs}
\begin{equation}
L_m^{r^B}(S_i) =\frac{1}{W}\sum_{S_{j}\in \mathbf{B} _i^{m,r^B}} \omega(S_i,S_j)   , \label{ms}
\end{equation}}%
\noindent where $\mathbf{B} _i^{m,r^B}$ is the set of matches for $S_i$ computed at scale $r^B$ and having a ground truth label $m$,
and $W$ is the normalization factor $W={\sum_{m=1}^M\sum_{S_j\in \mathbf{B} _i^{m,r^B}} \omega(S_i,S_{j})}$, with $\omega$ a weight depending on the DSP similarity \cite{giraud2017_spm}.
The final label of a superpixel $S_i$ is computed as
  $\mathcal{L} (S_i ) =
  \underset{m\in {\{1,\dots ,M\}}}{\argmax}\left(\underset{r^B\in \mathbf{r^B}}{\argmax}\left(L_m^{r^B}(S_i)\right)\right)$.

{\color{modifs}
In \fig~\ref{fig:expe_ms}, we represent matching results obtained
without (gray lines) and with (green lines) our multi-scale strategy.
}%
We can see that the best match between searches at scales $\mathbf{r^B}=[0.5, 1, 2, 4] {\times} r^A$ enables to catch the larger flower with similar colors, instead of the one at the same scale.

\section{Experimental Validations}
\label{sec:experiments}

In this section, we present several quantitative experiments to demonstrate the interest of the proposed DSPM framework.
We first validate the behavior of our model on standard images with respect to the method parameters.
Then, we propose larger scale segmentation experiments on a standard face dataset.

\subsection{Parameter Settings}

The proposed method was implemented with MATLAB using
C-MEX code on a standard Linux computer with $4$ cores at $1.90$ GHz and $16$ GB of RAM.
The number of DSPM iterations is set to 5,
and we use a $\ell_2$-norm as distance $d$ between features $F$ as in \cite{giraud2017_spm}.
{\color{v2}To avoid over detection, interfaces
are detected at least each
$4$ pixels.
}
Default parameters are set such that $\beta=1$, the border offset for region features $R$,
$\alpha=0.5$, the trade-off parameter between intra-region and interface distances in \eq~\eqref{full_model},
and superpatch radius $r=50$.
The considered descriptors in $R$ and $I$ are reported according to the application.

\subsection{Influence of Parameters}

To demonstrate the interest of each contribution,
{\color{v2}
we consider a matching experiment on standard images \textit{Baboon, Barbara, House, Lena} and \textit{Peppers},} %
each decomposed with two superpixel methods: SLIC~\cite{achanta2012} and SNIC~\cite{achanta2017snic}.
For each superpixel in a given decomposition,
we compute the closest DSP match in the other one.
A robust descriptor should indeed be robust to variations in the segmentations.
{\color{v2}
In terms of features, we compute
normalized cumulative RGB histogram with $9$ bins per canal on intra-regions $F_{R_{I'}}$, and HoG~\cite{dalal2005}
on a local $9{\times}9$ pixels window for interface region descriptors $F_{I_{I'}}$.}

\begin{figure}[t]
{\scriptsize
\begin{tabular}{@{\hspace{0mm}}c@{\hspace{1mm}}c@{\hspace{0mm}}}
\includegraphics[width=0.482\linewidth,height=0.16\textwidth]{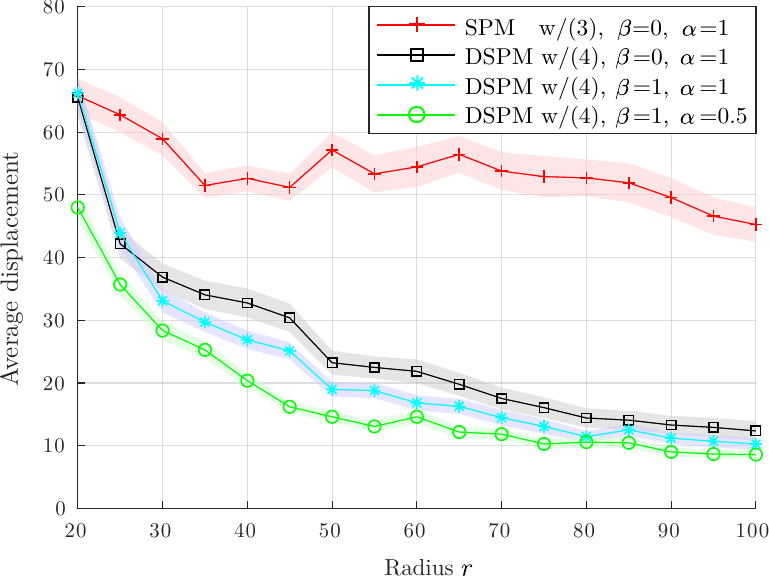}&
\includegraphics[width=0.482\linewidth,height=0.16\textwidth]{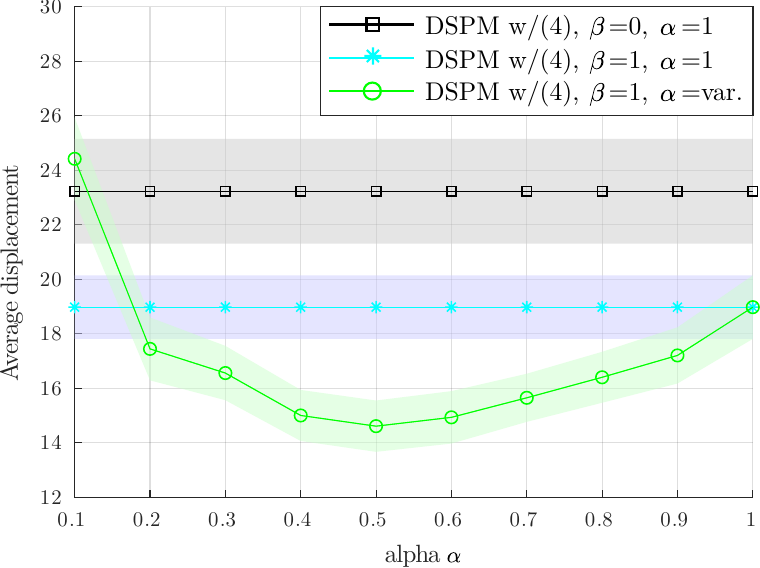}\\
(a)&(b)
\end{tabular}}

\caption{Influence of radius $r$ (a) and trade-off coefficient $\alpha$ (b) in the proposed DSPM framework
{\color{modifs}
compared to SPM.
}%
We report the average distance between each superpixel barycenter and the one of its closest match in another decomposition of the same image.
{\color{modifs}
See text for more details.
}%
}
\label{fig:expe_5}
\end{figure}

We evaluate the matching accuracy by the average distance between the superpixel barycenters and the one of their match in the other decomposition.
In \fig~\ref{fig:expe_5}(a) and \fig~\ref{fig:expe_5}(b),
we respectively report the average distance with respect to the radius parameter $r$
and with respect to the $\alpha$ parameter for $r=50$.
On the first hand, \fig~\ref{fig:expe_5}(a) shows that the accuracy logically increases with the superpatch radius,
and with each contribution, \emph{i.e.}, symmetrical projected distance (\eq~\eqref{dist_R}), offset to border ($\beta>0$), and interface descriptors.
On the other hand, \fig~\ref{fig:expe_5}(b) illustrates the interest of using interface descriptors in \eq~\eqref{full_model} in conjunction with cropped regions for
$\alpha > 0.2$, and that
a balanced trade-off parameter $\alpha\approx0.5$ provides the best matching accuracy.
In \fig~\ref{fig:expe_5_map}, we also show an example of matching result for DSPM with default parameters compared to SPM.

\newcommand\sih{0.07\textwidth}
\newcommand\sihh{0.115\textwidth}
\begin{figure}[t]
{\scriptsize
\begin{tabular}{@{\hspace{0mm}}c@{\hspace{1mm}}c@{\hspace{2mm}}c@{\hspace{1mm}}c}
\includegraphics[width=\sihh]{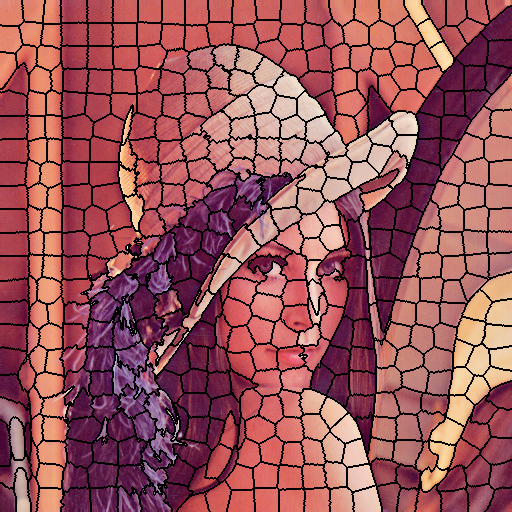}&
\includegraphics[width=\sihh]{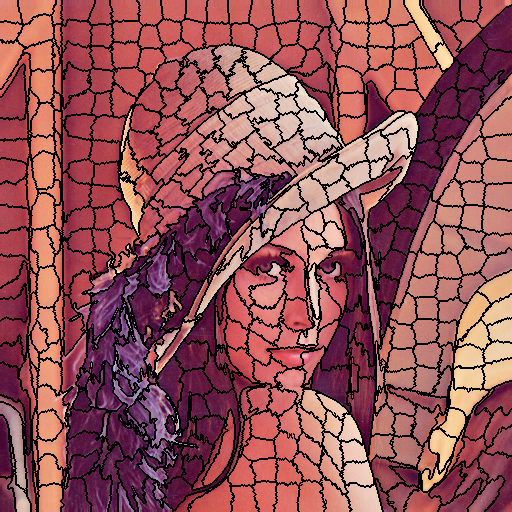}&
\includegraphics[width=\sihh]{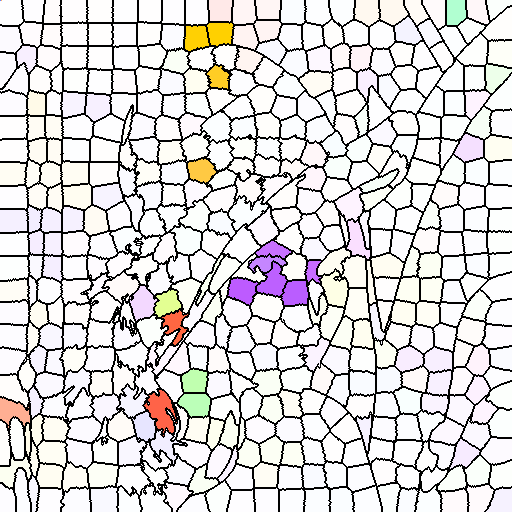}&
\includegraphics[width=\sihh]{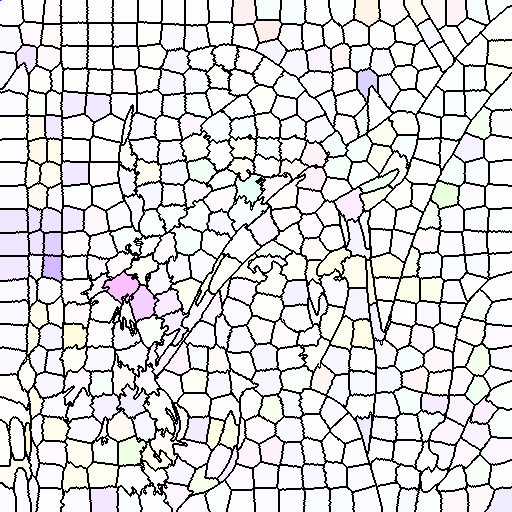}\\
(a) SLIC & (b) SNIC & (c) SPM & (d) DSPM
\end{tabular}
}%
\caption{Matching accuracy of DSPM vs SPM.
(a) and (b): decompositions with SLIC and SNIC.
(c) and (d): SPM
and DSPM matching results ($r$=$50$).
Distances between superpixel barycenters are shown with standard optical flow representation (stronger colors represent larger displacements).}
\label{fig:expe_5_map}
\end{figure}

\begin{table}[t]
\renewcommand{\arraystretch}{1.1}
\caption{Labeling accuracy for the multi-scale experiment.
Training images have been either downsampled or upsampled by a factor of 1.5 or 2 {\color{modifs}and $r^A$ is set to $50$}.
The argmax columns correspond to the fusion strategy proposed in \sect~\ref{subsec:fusion} for different combined scales.}
\begin{center}  
{\footnotesize
\begin{tabular}{@{\hspace{0mm}}c@{\hspace{3mm}}c@{\hspace{1.5mm}}c@{\hspace{1.5mm}}c@{\hspace{1.5mm}}c@{\hspace{1.5mm}}c@{\hspace{3mm}}c@{\hspace{1.5mm}}c@{\hspace{0mm}}}
\hline
\multirow{1}{*}{Radius $r^B$} & \multirow{1}{*}{50$^{*}$} & \multirow{1}{*}{25$^{*}$}& \multirow{1}{*}{33$^{*}$} & \multirow{1}{*}{75$^{*+}$}& \multirow{1}{*}{100$^{*+}$} & argmax$^*$  & argmax$^+$\\
w/o \eqref{rescale} & ${94.08\%}$ & $94.05\%$ & $93.93\%$ & $94.11\%$ & $94.07\%$ & $94.25\%$ & $94.16\%$ \\
w/ \eqref{rescale} & ${94.08\%}$ & $94.22\%$ & $94.13\%$ & $94.87\%$ &  $94.80\%$ & $94.78\%$ & {$\mathbf{94.96\%}$} \\
\hline
\end{tabular}}%
\end{center}
\label{table:ms}
\end{table}

\subsection{\label{sec:expe_lfw}Segmentation and Labeling Experiments}

\textit{Validation framework.}
We evaluate the proposed DSPM approach on the same face labeling experiment than \cite{giraud2017_spm}.
The considered Labeled Faces in the Wild (LFW) dataset \cite{huang2007db},
contains 1500 training and 927 testing images of $250{\times}250$ pixels,
linearly registered with
 \cite{huang2007},
and already decomposed into approximately 250 superpixels.
LFW contains decompositions and labeling ground truths,
so comparisons with state-of-the-art methods do not depend on the superpixel decompositions.
Note that
to fairly compare with \cite{giraud2017_spm},
we use the same HoG implementation on a regular grid
\cite{lsvm-pami} and compute 50 DSP matches by 50 independent DSPM processes for each superpixel, merged in Eq. \eqref{ms}.
  
\smallskip

\textit{Multi-scale validation.}
In this experiment, the goal is to validate the interest of our proposed multi-scale matching strategy introduced in Sec.~\ref{sec:multiscale}.
To this end, we have manually applied random scaling transformations to the registered training images.
Each image and its corresponding decomposition has been either downsampled or upsampled randomly by a factor of $1.5$ or $2$ (with no interpolation).
As a result, faces depicted in the images can appear up to twice as big or small compared to their initial scales.
{\color{v2}Hence, the dataset contains face patterns at different scales that would not be efficiently captured using the same DSP radius in $A$ and $B$.}

We apply DSPM with $r^A$=$50$ and $r^B\in \{25, 33, 50, 75, 100 \}$. 
In \tab~\ref{table:ms}, we report the labeling accuracy for each radius and for the multi-scale label fusion proposed in \sect~\ref{subsec:fusion}.
From these results, {\color{v2}when $r^A \neq r^B$}, we can observe that the performance for smaller or larger radiuses is always better after applying the rescale strategy, \emph{i.e.}, with \eq~\eqref{rescale}.
{\color{v2}The multi-scale fusion averaging the result over the different radius sizes performs better than the default $r^B = 50$ and than most single scales. 
Moreover, by considering only larger scales, \emph{i.e.}, $r^B = 75,100$, thus more precise DSP comparisons, we obtain the highest labeling accuracy, demonstrating the framework ability of the framework to merge information from multi-scale matching.}
In future works, we plan to apply this approach on non-registered datasets where objects might naturally appear at different scales.

\smallskip

\textit{Comparison to state-of-the-art methods.}
{\color{v2}
In \fig~\ref{fig:k_ann}, we first compare the influence of the number $k$ of selected ANN for each superpixel,
then merged in the label fusion process \eq~\eqref{ms} for the proposed DSPM method and
SPM \cite{giraud2017_spm} using \eq~\eqref{dist_R1}.
For all ANN numbers, DSPM provides the best results, and is already reaching $95.13\%$ of accuracy with only $k=20$ ANN.
Note that for both methods, a plateau is reached around $k=50$ ANN.

\begin{figure}[t]
\centering
\includegraphics[width=0.365\textwidth]{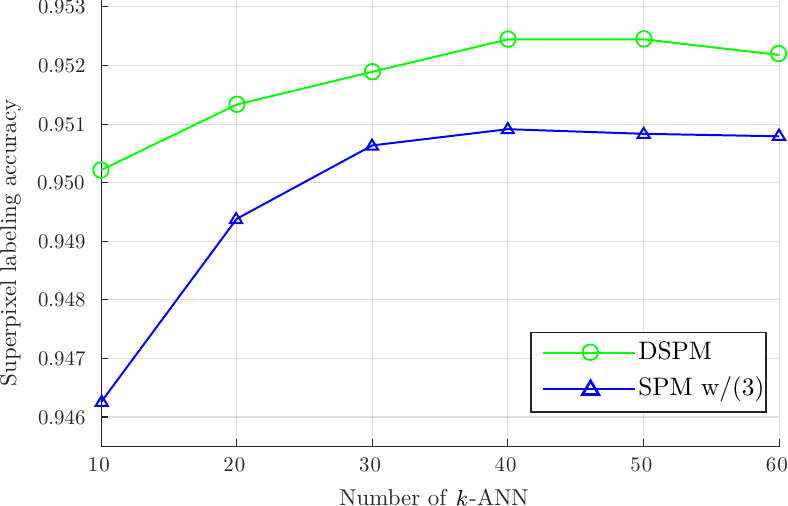}
\caption{Influence of the number of selected $k$-ANN for SPM \cite{giraud2017_spm} and DSPM
on face labeling accuracy.}
\label{fig:k_ann}
\end{figure}
}

\begin{table}[t]
\renewcommand{\arraystretch}{1.1}
\caption{Exemplar-based labeling accuracy on the LFW dataset.
}
\begin{center}
\newcommand{\sz}{\hspace{1pt}}
\newcommand{\szz}{\hspace{1pt}}
{\footnotesize
\begin{tabular}{@{\szz}p{3.9cm}@{\sz}c@{\hspace{5mm}}c@{\szz}}
\hline
Method & Superpixel accuracy & Pixel  accuracy\\
\hline
Spatial CRF \cite{kae2013}&$93.95\%$ &\textit{not reported}\\
CRBM \cite{kae2013}& $94.10\%$  &\textit{not reported}\\
GLOC \cite{kae2013}& $94.95\%$  &\textit{not reported}\\
DCNN \cite{liu2015}&\textit{not reported}& $95.24\%$ \\
SPM \cite{giraud2017_spm} w/ \eqref{dist3_spp}&${91.88\%}$ &${92.21\%}$\\
SPM \cite{giraud2017_spm} w/ \eqref{dist_R1}&${95.08\%}$ &${95.43\%}$\\
\textbf{DSPM}&{\color{v2}$\mathbf{95.24\%}$}&{\color{v2}$\mathbf{95.59\%}$}\\
\hline
\end{tabular}}
\label{table:LFW}
\end{center}
\end{table}

\begin{figure}[t!]
\newcommand{\sss}{0.09\textwidth}
\newcommand{\ssss}{0.0875\textwidth}
\centering
{\scriptsize
\begin{tabular}{@{\hspace{0mm}}c@{\hspace{1mm}}c@{\hspace{1mm}}c@{\hspace{1mm}}c@{\hspace{1mm}}c@{\hspace{0mm}}}
&&& $98.82\%$ & $100.00\%$ \\
\includegraphics[width=\sss,height=\ssss]{\pathresspm/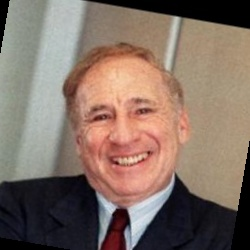} &
\includegraphics[width=\sss,height=\ssss]{\pathresspm/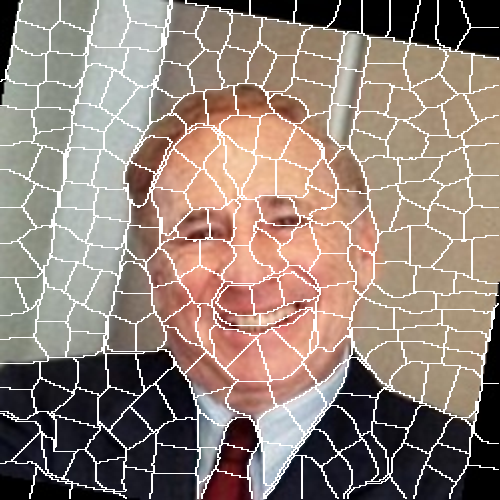} &
\includegraphics[width=\sss,height=\ssss]{\pathresspm/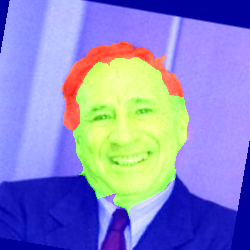} &
\includegraphics[width=\sss,height=\ssss]{\pathresspm/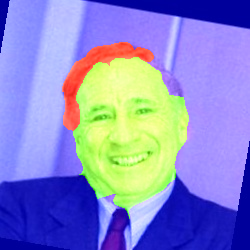} &
\includegraphics[width=\sss,height=\ssss]{\pathresspm/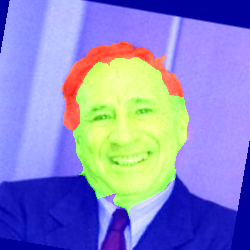} \\
&&& $90.30\%$ & $97.76\%$ \\
\includegraphics[width=\sss,height=\ssss]{\pathresspm/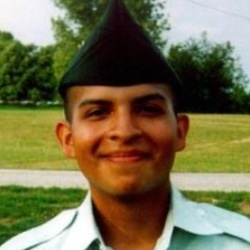} &
\includegraphics[width=\sss,height=\ssss]{\pathresspm/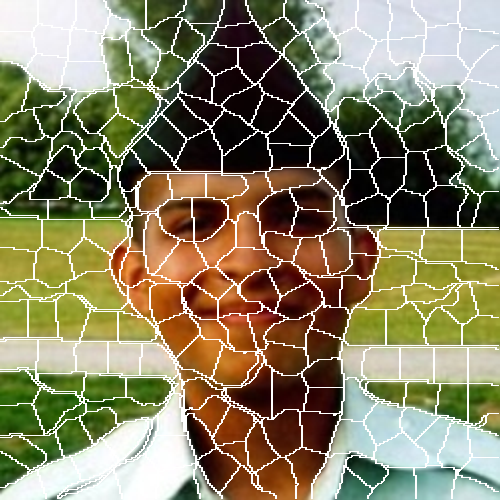} &
\includegraphics[width=\sss,height=\ssss]{\pathresspm/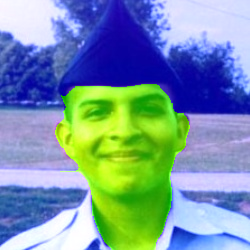} &
\includegraphics[width=\sss,height=\ssss]{\pathresspm/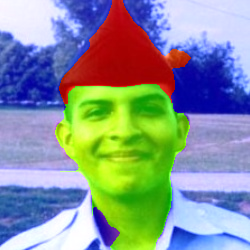} &
\includegraphics[width=\sss,height=\ssss]{\pathresspm/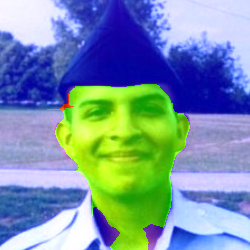} \\
&&& $94.21\%$ & $97.68\%$ \\
\includegraphics[width=\sss,height=\ssss]{\pathresspm/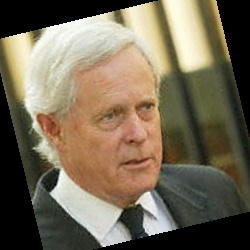} &
\includegraphics[width=\sss,height=\ssss]{\pathresspm/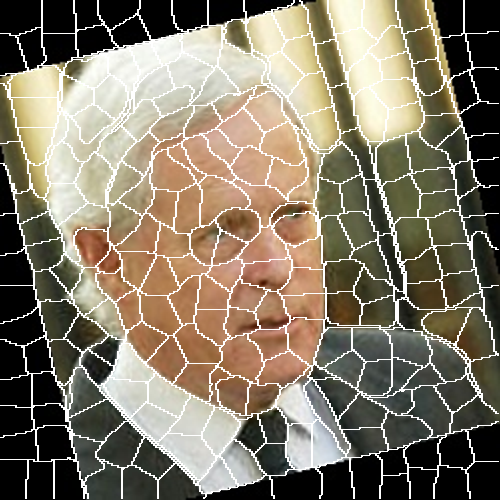} &
\includegraphics[width=\sss,height=\ssss]{\pathresspm/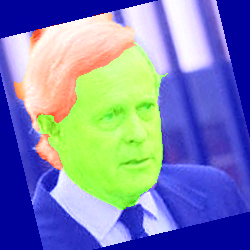} &
\includegraphics[width=\sss,height=\ssss]{\pathresspm/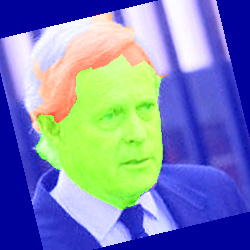} &
\includegraphics[width=\sss,height=\ssss]{\pathresspm/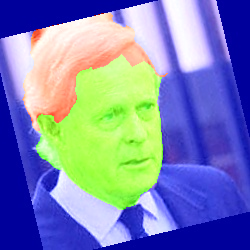} \\
&&& $92.25\%$ & $96.90\%$ \\
\includegraphics[width=\sss,height=\ssss]{\pathresspm/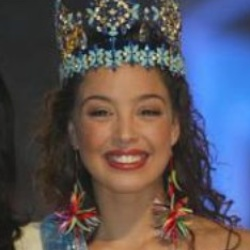} &
\includegraphics[width=\sss,height=\ssss]{\pathresspm/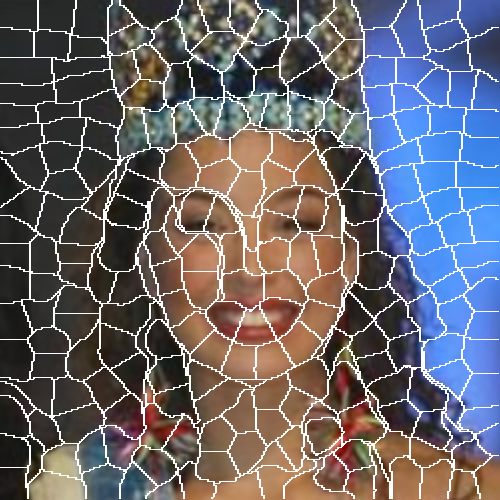} &
\includegraphics[width=\sss,height=\ssss]{\pathresspm/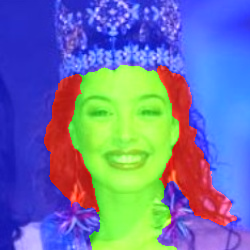} &
\includegraphics[width=\sss,height=\ssss]{\pathresspm/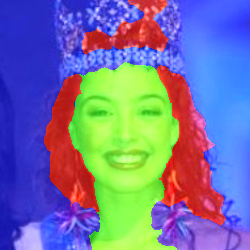} &
\includegraphics[width=\sss,height=\ssss]{\pathresspm/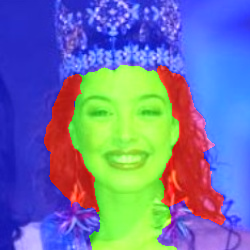} \\
Image & Superpixels & Ground truth & SPM & \textbf{DSPM} \\ 
\end{tabular}}
\caption{Examples of labeling results with superpixel accuracy obtained with the proposed DSPM approach on the LFW dataset, compared to the initial SPM method.} 
\label{fig:lfw_results}
\end{figure}

In \tab~\ref{table:LFW}, we also compare the performance of the proposed DSPM method with the results of state-of-the-art ones, mostly based on supervised (deep) learning approaches.
{\color{v2}
In \cite{kae2013}, several approaches are used to label the LFW dataset such as
a spatial conditional random field (CRF) and a conditionnal restricted Boltzmann machine (CRBM).
The GLOC (GLObal and LOCal) method \cite{kae2013} is also proposed to jointly use both CRF and CRNM approaches
to introduce global shape priors in the training process.
Finally, in \cite{liu2015}, a deep convolutional neural
network (DCNN) is proposed and dedicated to the face labeling application.
Note that the results in Table \ref{table:LFW} are the ones reported by the authors.}
For SPM, the results correspond to the initial framework using costly quadratic comparisons with \eq~\eqref{dist3_spp}, and the results reported by the authors using non-symmetric projected distances with \eq~\eqref{dist_R1}.
DSPM reports the best compared labeling accuracy at both superpixel and pixel-wise level.
Labeling examples compared to SPM are also represented in \fig~\ref{fig:lfw_results}.
The proposed DSP enables to relevantly capture the context of a superpixel neighborhood in terms of texture and structure.
Moreover, without further optimizations on non fully multi-threaded code, DSPM performs in less than $3$s per subject, against $45$s for SPM using \eq~\eqref{dist3_spp}.
Note that compared state-of-the-art approaches may provide faster computational times but at the expense of previous costly learning-based processes.

Our method is particularly interesting due to its simplicity of use, parameter tuning,
and interpretability compared to learning-based approaches, while providing more accurate results than SPM.
{\color{v2}Besides, any feature can be directly used in the method, even more advanced descriptors, \emph{e.g.}, \cite{zhang2018consistent,tilquin2018robust}, eventually based on previously trained deep learning architectures.}

\section{Conclusion}

In this work, we addressed some important limitations of existing superpixel matching frameworks,
in terms of robustness and computational complexity.
We introduced the dual superpatch, a new superpixel neighborhood descriptor containing both
{\color{modifs}
intra-region and interface information
that are respectively
robust to the inaccuracy of superpixel borders and capture contour structures.
}%
We also proposed optimized distances and a multi-scale framework to search for similar dual superpatches in an image dataset.
Our validations showed an accuracy improvement with our method on matching and exemplar-based labeling applications.
{\color{modifs}
The relevance of the proposed dual approach should benefit to all superpixel-based non-local approaches and
}%
future works will focus on applying the method to heterogeneous computer vision and medical datasets,
and tackling other
applications such as superpixel-based image editing.

\bibliographystyle{model2-names}
\bibliography{paper}

\end{document}